\documentclass{article}

\usepackage[preprint]{neurips_2021}

\usepackage[utf8]{inputenc} % allow utf-8 input
\usepackage[T1]{fontenc}    % use 8-bit T1 fonts
\usepackage{url}            % simple URL typesetting
\usepackage{booktabs}       % professional-quality tables
\usepackage{amsfonts}       % blackboard math symbols
\usepackage{nicefrac}       % compact symbols for 1/2, etc.
\usepackage{microtype}      % microtypography
\usepackage{xcolor}         % colors
\usepackage{graphicx}

\title{InfiniteForm: A synthetic, minimal bias dataset for fitness applications}

\author{%
  Andrew Weitz, PhD \\
  Edge Analytics Inc.\\
  Campbell, CA \\
  \texttt{andrew@edgeanalytics.io} \\
   \And
   Lina Colucci, PhD \\
   Edge Analytics Inc.\\
   Campbell, CA \\
   \texttt{lina@edgeanalytics.io} \\
   \AND
   Sidney Primas, MS \\
   Edge Analytics Inc.\\
   Campbell, CA \\
   \texttt{sidney@edgeanalytics.io} \\
   \And
   Brinnae Bent, PhD \\
   Edge Analytics Inc.\\
   Campbell, CA \\
   \texttt{brinnae@edgeanalytics.io} \\
}

\begin{document}

\maketitle

\begin{abstract}
The growing popularity of remote fitness has increased the demand for highly accurate computer vision models that track human poses. However, the best methods still fail in many real-world fitness scenarios, suggesting that there is a domain gap between current datasets and real-world fitness data. To enable the field to address fitness-specific vision problems, we created InfiniteForm \textendash{} an open-source synthetic dataset of 60k images with diverse fitness poses (15 categories), both single- and multi-person scenes, and realistic variation in lighting, camera angles, and occlusions. As a synthetic dataset, InfiniteForm offers minimal bias in body shape and skin tone, and provides pixel-perfect labels for standard annotations like 2D keypoints, as well as those that are difficult or impossible for humans to produce like depth and occlusion. In addition, we introduce a novel generative procedure for creating diverse synthetic poses from predefined exercise categories. This generative process can be extended to any application where pose diversity is needed to train robust computer vision models.

\end{abstract}

\section{Introduction}

Remote fitness has seen tremendous growth in recent years, ushered in by companies like Peloton [1] and Tempo [2], a variety of exercise apps like Apple Fitness+ [3], and social distancing protocols introduced during the COVID-19 pandemic. This has increased the demand for real-time computer vision systems that can accurately analyze the human form. Such models can be used to give users personalized feedback during at-home workouts, including correction of exercise posture [4], estimation of energy expenditure [5], and exercise repetition counting [6]. Unfortunately, the best vision methods still fail in many real-world fitness scenarios [7] suggesting that there is a domain gap between current datasets and real-world fitness scenes. 

Building a dataset for remote fitness has specific needs, including a diversity of complex poses, variation in viewpoints, and challenging lighting conditions. To our knowledge there are no domain-specific, open-source datasets for remote fitness that fit these needs. Datasets like COCO [8], MPII [9], SURREAL [10], DensePose [11], and AGORA [12] have been critical to the success of general purpose pose estimation, but do not offer breadth in the poses most relevant to remote fitness. The Leeds Sports Pose dataset [13] was the first of its kind to target exercise, but is relatively small (2k images) and only includes one type of annotation. Similarly, the Yoga-82 dataset [14] (28.4k images) covers some of the most diverse poses that a human body can perform, but only includes semantic-level annotations like pose name and category.

To address these needs, we created InfiniteForm \textendash{} an open source, synthetic dataset that is specific to the  fitness domain. The dataset consists of 60k images with both single- and multi-person scenes (up to 5 people) where each person is doing unique variations of 15 different fitness pose categories. The dataset offers a rich set of annotations \textendash{} including pose classification, 2D and 3D keypoints, occluded joints, semantic segmentation, depth, and more \textendash{} to enable a wide variety of vision models to be both evaluated on and trained on this data. As a synthetic dataset, InfiniteForm was explicitly designed to minimize biases in gender, skin tone, and body shape, which often exist in human-collected, real-world data [15] and can have severe consequences for resulting ML models [16-20].

\section{Dataset}
\label{gen_inst}

\subsection{Synthetic Data Generation}

\begin{figure}
  \centering
  \includegraphics[width=5.5in]{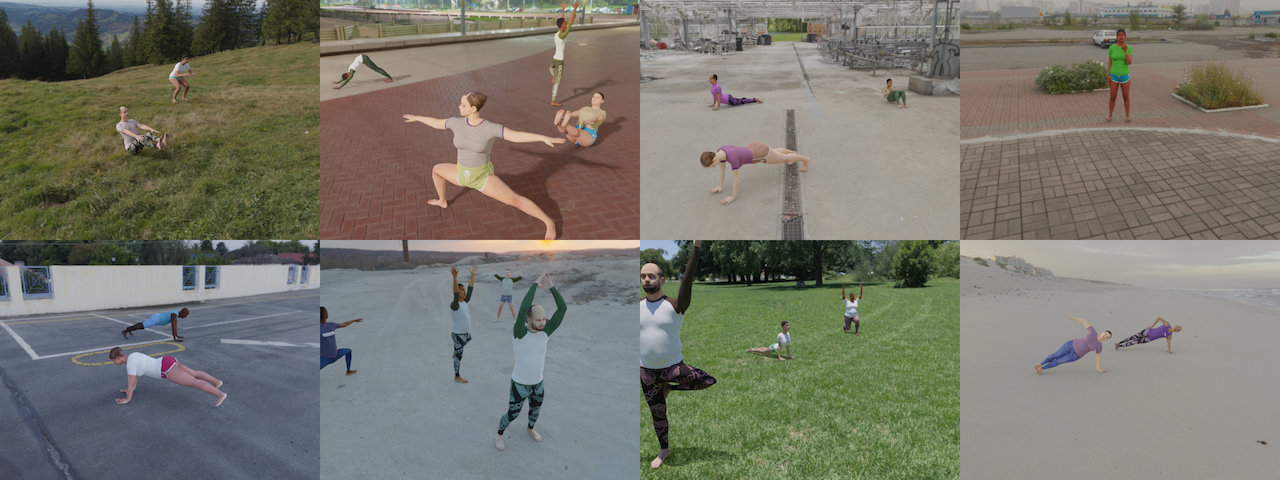}
  \caption{Example images rendered in the InfiniteForm dataset. Each avatar is positioned in a full body pose designed to unambiguously fall into a specific exercise.}
\end{figure}

The InfiniteForm dataset consists of 60k (640x480)-resolution images, each including 1-5 realistic avatars (Figure 1) in a variety of annotated, fitness-specific poses. We used the open-source software Blender [21] and a physics-based rendering engine with ray-tracing to maximize photorealism. The SMPL-X body model [22] was used for each avatar. To introduce realistic variation of body dimensions, the SMPL-X shape and facial expression parameters were randomly sampled from standard normal and [-2,2) uniform distributions, respectively. Clothing and skin tone were also randomly selected from a base library of 50 4K-resolution UV textures, designed to reflect global demographics for ethnicity and spanning seven unique garments [23]. Additional variation in skin tone and clothing was introduced by independently and randomly perturbing the RGB curves of each.

Varied and natural lighting was achieved by using 147 4K HDRI panoramas as the image background and lighting source [24]. This technique avoids composition artifacts introduced in other human pose datasets when avatars are overlaid onto random background images, without accounting for environmental lighting. A transparent ground plane was used to achieve realistic shadows. A mask is provided to facilitate replacing the backgrounds with additional images. 

Avatars were placed at random distances from the camera with a minimum distance of 2 meters between each other, and randomly rotated along the z-axis between -90 and +90 degrees. The camera was positioned at random elevations in a radial coordinate system to achieve a diversity of camera perspectives. After positioning, the camera was rotated so that the average location of all avatars in the image was mapped to the image center. This scene configuration resulted in diverse occlusion and self-occlusion patterns. A fixed focal length of 35mm was used for all images.

\subsection{Synthetic Pose Generation}

Avatars in the dataset are positioned in full body poses designed to unambiguously fall into a specific exercise category. The dataset includes 15 pose categories with 21 pose variations (Figure 3). For example, “lunge” is a pose category in the dataset and “left leg lunge” is a pose variation. For each pose variation, we employ one of two processes to create a diverse set of unique poses that are randomly sampled at the time of dataset creation. We note that “pose” here and throughout the paper refers to a unique set of joint rotations in the SMPL-X body model.

A novel generative process was designed to import poses from the real world and add subtle variations while maintaining biomechanical feasibility. We start by recording a small number of videos in which a subject moves through various examples of a particular pose variation for 10-40 seconds, and using a learned model to estimate 3D poses from each frame [25]. Next, we map the extracted poses to a 32-dimensional, normally distributed latent space that encodes biomechanically feasible poses [22], and fit a multivariate normal distribution to each pose variation’s embeddings. Finally, we sample new pose embeddings from the learned distribution and decode them into entirely new 3D poses that maintain the qualitative essence of the original source videos. We believe the usefulness of this pipeline extends beyond the current dataset, by providing a cost-effective and scalable process for auto-importing 3D poses from the real-world via RGB images or video, and adding realistic variation. 

For pose categories in which precise articulation of joints or ground-plane interactions are required to maintain photorealism, a small set of keyframe poses were manually designed in Blender. Bezier curve interpolations were then performed to generate new poses between keyframe pose pairs. 

\begin{figure}
  \centering
  \includegraphics[width=5.5in]{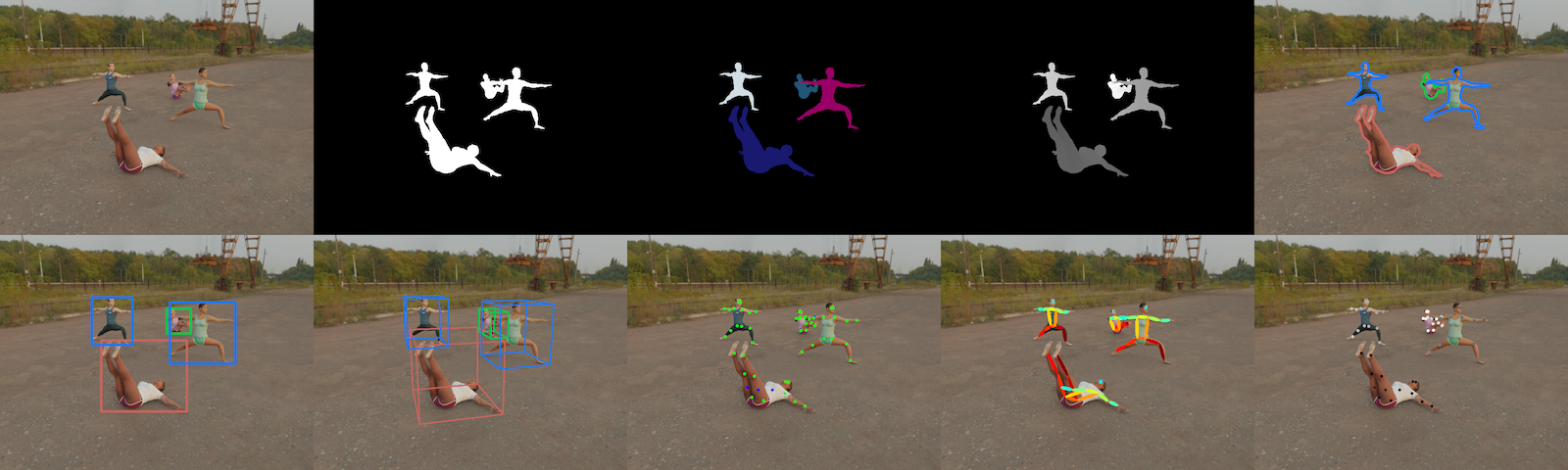}
  \caption{ Labels provided in the InfiniteForm dataset for an example image. Top row: Original RGB image, semantic segmentation, instance segmentation, 32-bit depth map, and polygon segmentation. Bottom row: 2D bounding boxes, 3D cuboids, 2D keypoint visibility (green=visible; blue=occluded), 2D keypoint skeleton, 3D keypoints. Depth maps and 3D keypoints are normalized for visualization. Colors of polygon, bounding box, and cuboid annotations represent pose category.}
\end{figure}

\subsection{Provided Labels}

A major advantage of synthetic datasets is having access to labels that are either difficult for humans to annotate, like polygon segmentation, or impossible, like depth maps and occluded keypoints. For each image in the InfiniteForm dataset, we provide pixel-level semantic segmentation, instance segmentation, and 32-bit OpenEXR images that encode unnormalized depth from the camera. For each, the SMPL-X avatars are the only objects labeled in the scene. 2D bounding boxes, polygon segmentation, 2D keypoints, 3D keypoints, and 3D cuboids (in image space) are also provided for each avatar (Figure 2). Annotations were organized using the ZPY open-source framework [26].

A major advantage of synthetic datasets for pose estimation is that ground-truth labels can be provided for all keypoints, regardless of occlusion status. Following standard COCO format, we annotate whether keypoints are either (a) not present in the image, (b) present in the image but hidden by another avatar or self-occlusion, or (c) visible. A keypoint is considered to be hidden by self-occlusion if a ray cast from the camera to the keypoint hits the body more than 0.3 meters from the keypoint.

Additional image and avatar-specific metadata provided in the dataset includes: camera pitch, HDRI background and rotation, pose category and variation, avatar yaw angle, body shape (SMPL-X shape coefficients, circumference of waist, and height), presenting gender, and clothing.

\begin{figure}
  \centering
  \includegraphics[width=5.5in]{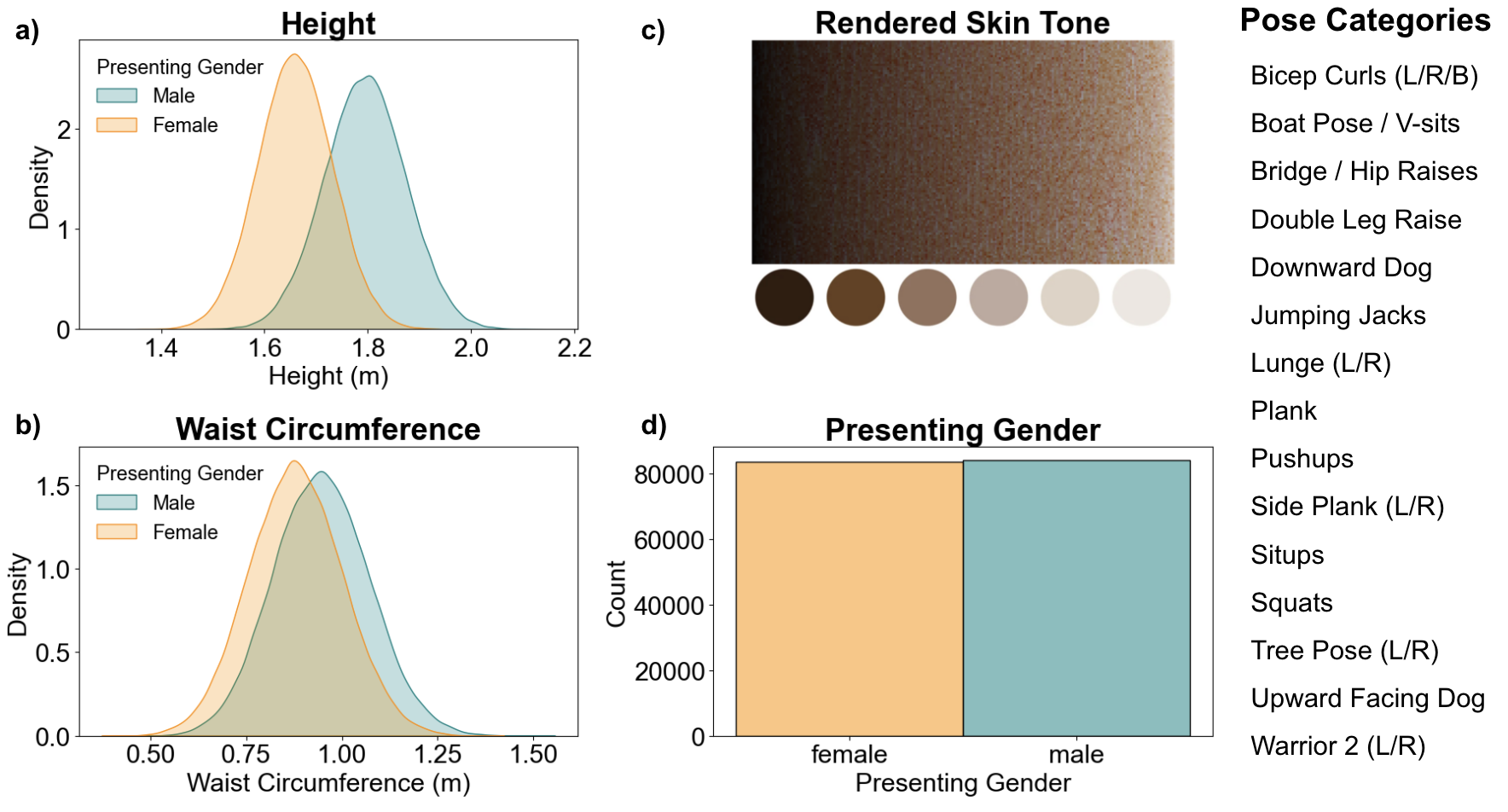}
  \caption{ InfiniteForm Dataset Descriptive Statistics. (a) Height is normally distributed in the dataset for each presenting gender. (b) Waist circumference is normally distributed in the dataset. (c) Rendered skin tones were extracted between the eye keypoints and are visualized here, sorted based on the standard HSV color model representation value (V) of lightness/brightness. Shown below the distribution of skin tones is the standard Fitzpatrick skin tone scale [29]. (d) Presenting gender (male/female) are equally represented in the dataset. Additionally, the 15 pose categories are shown.}
\end{figure}

\subsection{Descriptive Statistics and Limitations}

Distributions of rendered skin tone, height, waist circumference, and presenting gender are shown in Figure 3. Waist circumference is normally distributed with an average value of 0.91 meters. This value lies between the recommended average for waist circumference [27] and the United States population average [28]. Height is normally distributed, with distinct distributions for each of the presenting genders with a dataset average of 1.73 meters. Each of the skin tones on the standard Fitzpatrick skin tone scale [29] are represented in the dataset (Figure 3). 

The dataset has several limitations, including a lack of: diversity in age; complex hair, clothing, and footwear; fitness-related equipment like dumbbells; and other accessories, such as headscarves and glasses. In addition, the adoption of equirectangular HDRIs as background images sometimes results in unrealistic composition artifacts in which avatars are not placed on the visible ground plane. The dataset consists of only 15 pose categories, which will likely need future expansion to provide robust coverage across all fitness applications. Despite these limitations, we believe InfiniteForm provides a much needed resource to those working on vision problems in the fitness domain.

\section{Dataset Availability}

InfiniteForm is licensed under a Creative Commons Attribution 4.0 License. To download the dataset and learn more about InfiniteForm, please visit: \url{https://pixelate.ai/InfiniteForm}.

\section*{References}

\medskip

{
\small

[1] “Peloton | Workouts Streamed Live and On-Demand.” [Online]. Available: https://www.onepeloton.com/. [Accessed: 30-Sep-2021].

[2] “The Award-Winning AI-Powered Home Gym | Tempo.” [Online]. Available: https://tempo.fit/. [Accessed: 30-Sep-2021].

[3] “Apple Fitness+ | Apple.” [Online]. Available: https://www.apple.com/apple-fitness-plus/. [Accessed: 30-Sep-2021].

[4] S. Chen, R. Yang, “Pose Trainer: Correcting Exercise Posture using Pose Estimation”. Jun. 2020. https://arxiv.org/abs/2006.11718

[5] P. Saponaro, H. Wei, G. Dominick, and C. Kambhamettu, “Estimating Physical Activity Intensity And Energy Expenditure Using Computer Vision On Videos,” IEEE Int. Conf. on Image Processing (ICIP) 2019.  

[6] T. Alatiah and C. Chen, “Recognizing Exercises and Counting Repetitions in Real Time,” 2020. https://arxiv.org/pdf/2005.03194.pdf

[7] A. Garbett, Z. Degutyte, J. Hodge, and A. Astell, “Towards Understanding People’s Experiences of AI Computer Vision Fitness Instructor Apps,” DIS 2021 - Proc. 2021 ACM Des. Interact. Syst. Conf. Nowhere Everywhere, pp. 1619–1637, Jun. 2021.

[8] T.Y. Lin et al., “Microsoft COCO: Common Objects in Context,” Lect. Notes Comput. Sci. (including Subser. Lect. Notes Artif. Intell. Lect. Notes Bioinformatics), vol. 8693 LNCS, no. PART 5, pp. 740–755, May 2014.

[9] M. Andriluka, L. Pishchulin, P. Gehler, and B. Schiele, “2D Human Pose Estimation: New Benchmark and State of the Art Analysis,” IEEE Conf. Comput. Vis. Pattern Recognit., 2014.

[10] G. Varol et al., “Learning from Synthetic Humans,” Proc. - 30th IEEE Conf. Comput. Vis. Pattern Recognition, CVPR 2017, vol. 2017-January, pp. 4627–4635, Jan. 2017.

[11] R. A. Güler, N. Neverova, and I. Kokkinos, “DensePose: Dense Human Pose Estimation In The Wild,” Proc. IEEE Comput. Soc. Conf. Comput. Vis. Pattern Recognit., pp. 7297–7306, Feb. 2018.

[12] P. Patel, C.H. P. Huang, J. Tesch, D. T. Hoffmann, S. Tripathi, and M. J. Black, “AGORA: Avatars in Geography Optimized for Regression Analysis,” Apr. 2021.

[13] S. Johnson and M. Everingham, “Clustered Pose and Nonlinear Appearance Models for Human Pose Estimation Mark Everingham,” in Proceedings of the 21st British Machine Vision Conference (BMVC2010), 2010.

[14] M. Verma, S. Kumawat, Y. Nakashima, and S. Raman, “Yoga-82: A New Dataset for Fine-grained Classification of Human Poses.” https://arxiv.org/pdf/2004.10362v1.pdf

[15] S. Alelyani, “Detection and Evaluation of Machine Learning Bias,” Appl. Sci. 2021, Vol. 11, Page 6271, vol. 11, no. 14, p. 6271, Jul. 2021.

[16] M. A. Gianfrancesco, S. Tamang, J. Yazdany, and G. Schmajuk, “Potential Biases in Machine Learning Algorithms Using Electronic Health Record Data,” JAMA Intern. Med., vol. 178, no. 11, p. 1544, Nov. 2018.

[17] Z. Obermeyer, B. Powers, C. Vogeli, and S. Mullainathan, “Dissecting racial bias in an algorithm used to manage the health of populations,” Science (80), vol. 366, no. 6464, pp. 447–453, Oct. 2019.

[18] S. Leavy, “Uncovering gender bias in newspaper coverage of Irish politicians using machine learning,” Digit. Scholarsh. Humanit., vol. 34, no. 1, pp. 48–63, Apr. 2019.

[19] M. O. R. Prates, P. H. Avelar, and L. C. Lamb, “Assessing gender bias in machine translation: a case study with Google Translate,” Neural Comput. Appl., vol. 32, no. 10, pp. 6363–6381, May 2020.

[20] J. Buolamwini and T. Gebru, "Gender shades: Intersectional accuracy disparities in commercial gender classification," in Conference on Fairness, Accountability and Transparency (PMLR), 2018.

[21] “blender.org - Home of the Blender project - Free and Open 3D Creation Software.” [Online]. Available: https://www.blender.org/. [Accessed: 30-Sep-2021].

[22] G. Pavlakos et al., “Expressive Body Capture: 3D Hands, Face, and Body from a Single Image,” Proc. IEEE Comput. Soc. Conf. Comput. Vis. Pattern Recognit., vol. 2019-June, pp. 10967–10977, Apr. 2019.

[23] “Meshcapade GmbH.” [Online]. Available: https://meshcapade.com/. [Accessed: 30-Sep-2021].

[24] “Poly Haven.” [Online]. Available: https://polyhaven.com/. [Accessed: 30-Sep-2021].

[25] M. Kocabas, N. Athanasiou, and M. J. Black, “VIBE: Video Inference for Human Body Pose and Shape Estimation,” Proc. IEEE Comput. Soc. Conf. Comput. Vis. Pattern Recognit., pp. 5252–5262, Dec. 2019.

[26] H. Ponte, N. Ponte, and S. Crowder, “zpy: Synthetic data for Blender,” GitHub, 2021. [Online]. Available: https://github.com/ZumoLabs/zpy. [Accessed: 30-Sep-2021].

[27] “Waist circumference is ‘vital sign’ of health, experts say.” [Online]. Available: https://www.medicalnewstoday.com/articles/waist-circumference-is-vital-sign-of-health-experts-say. [Accessed: 30-Sep-2021].

[28] “CDC FastStats - Body Measurements,” CDC. [Online]. Available: https://www.cdc.gov/nchs/fastats/body-measurements.htm. [Accessed: 30-Sep-2021].

[29] T. B. Fitzpatrick, “The validity and practicality of sun-reactive skin types I through VI,” Arch. Dermatol., vol. 124, no. 6, pp. 869–871, Jun. 1988.
}

\end{document}